\documentclass{sig-alternate-05-2015}
\usepackage{graphicx}
\usepackage{amsmath,amssymb}

\begin{document}






%

\title{Fashioning with Networks: Neural Style Transfer to Design Clothes}
%
%
%
%
%

\numberofauthors{3} 
%
\author{
%
%
\alignauthor
Prutha Date\\
       \affaddr{University Of Maryland Baltimore County (UMBC),}\\
       \affaddr{Baltimore, MD,}\\
       \affaddr{USA}\\
       \email{dprutha1@umbc.edu}
\alignauthor
Ashwinkumar Ganesan\\
       \affaddr{University Of Maryland Baltimore County (UMBC),}\\
       \affaddr{Baltimore, MD,}\\
       \affaddr{USA}\\
       \email{gashwin1@umbc.edu}
\alignauthor
Tim Oates\\
       \affaddr{University Of Maryland Baltimore County (UMBC),}\\
       \affaddr{Baltimore, MD,}\\
       \affaddr{USA}\\
       \email{oates@cs.umbc.edu}
}

\maketitle
\begin{abstract}
Convolutional Neural Networks have been highly successful in performing a host of computer vision tasks such as object recognition, object detection, image segmentation and texture synthesis. In 2015, Gatys \textit{et. al} \cite{gatys2015neural} show how the style of a painter can be extracted from an image of the painting and applied to another normal photograph, thus recreating the photo in the style of the painter. The method has been successfully applied to a wide range of images and has since spawned multiple applications and mobile apps. In this paper, the neural style transfer algorithm is applied to fashion so as to synthesize new custom clothes. We construct an approach to personalize and generate new custom clothes based on a user's preference and by learning the user's fashion choices from a limited set of clothes from their closet. The approach is evaluated by analyzing the generated images of clothes and how well they align with the user's fashion style.
\end{abstract}

%
%

\ccsdesc[500]{Computing methodologies~Computer vision}
\ccsdesc[300]{Computing methodologies~Machine learning approaches}
\ccsdesc[100]{Computing methodologies~Neural networks}

%
%

%
%
\printccsdesc


\keywords{Convolutional Neural Networks, Personalization, Fashion, Neural Networks, Style Transfer, Texture Synthesis}

\section{Introduction}
There have been recently impressive advances in computer vision tasks like object recognition and detection, segmentation \cite{krizhevsky2012imagenet}\cite{DBLP:journals/corr/RenHG015}\cite{DBLP:journals/corr/ChenPK0Y16}. The revolution started with Krizhevsky \textit{et. al} \cite{krizhevsky2012imagenet} substantially improving object recognition on the Imagenet challenge using convolutional neural networks (CNN). This led to research and subsequent improvements in many tasks related to fashion such as classification of clothes, predicting different kinds of attributes of a specific piece of clothing, and improving the retrieval of images \cite{liu2016deepfashion}\cite{kalantidis2013getting}\cite{bossard2012apparel}\cite{xiao2015learning}\cite{kiapour2014hipster}. Giants of e-commerce are expanding their investment in fashion. Recently, Amazon patented a system to manufacture clothes on demand \cite{AmazonFashion}. Also, they have started shipping their virtual assistant \textit{Echo} with an integrated camera that clicks a picture of the user's outfit and rates its \textit{style} \cite{AmazonCamera}. \textit{StitchFix}\footnote{https://www.stitchfix.com/welcome/schedule} aims to simplify the user's experience to shop online. As the online fashion industry looks to improve the kind of clothes that are recommended to users, understanding the personal style preferences of users and recommending custom designs becomes an important task. 

Personalization and recommendation models are a well researched area that include methods from collaborative filtering \cite{linden2003amazon} to content-based recommendation systems (e.g., probabilistic graph models, neural networks) as well as hybrid systems that combine both. Collaborative filtering \cite{linden2003amazon} tries to analyze user behaviour and preferences and align users to predefined patterns so as to recommend a product. Content-based methods recommend a product based on its attributes or features that the user is searching for. A hybrid system (knowledge-based system \cite{trewin2000knowledge}) incorporates user preferences and product features to recommend an item to the user.

\begin{figure*}[h]
\centering
\includegraphics[width=0.3\textwidth]{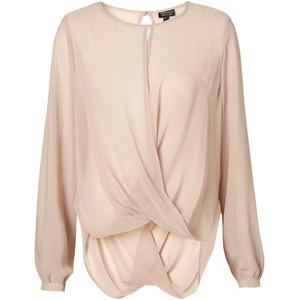}
\includegraphics[width=0.3\textwidth]{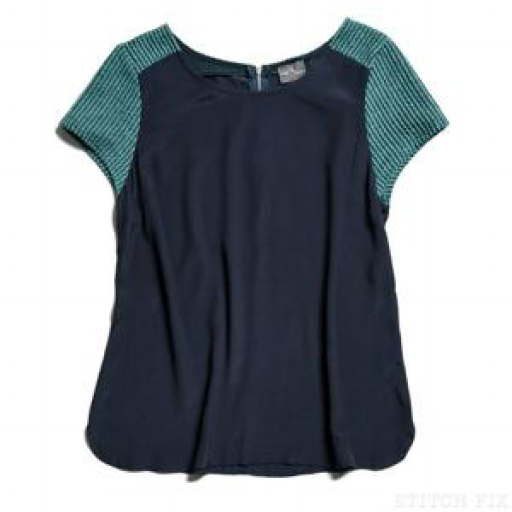}
\includegraphics[width=0.3\textwidth]{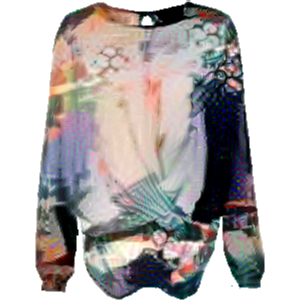}
\caption{(a) and (b) provide the shape \& style respectively (c) Final Design}
\label{fig:sample}
\end{figure*}

While the above techniques retrieve the product (or its image) we seek to synthesize new personalized merchandise. Texture synthesis tries to learn the underlying texture of an image in order to generate new samples with the same texture. The research in this space \cite{gatys2015texture} is largely focused on parametric and non-parametric methods. Non-parametric methods try to resample specific pixels from the image or adopt specific patches from the original to generate the new image \cite{efros1999texture}\cite{wei2000fast}\cite{kwatra2003graphcut}\cite{efros2001image}. Parametric methods define a statistical model that represents the texture \cite{julesz1962visual}\cite{heeger1995pyramid}\cite{portilla2000parametric}. In 2015, Gatys \textit{et. al}. \cite{gatys2015texture} designed a new parametric model for texture synthesis based on convolutional neural networks. They model the style of an image by extracting the feature maps generated when the image is fed through a pre-trained CNN, in this instance using a 19 layer VGGNet. They successfully separate the style and content of an arbitrary image and demonstrate how the other image can be stylized using the textures of the prior.

\begin{figure*}[h]
\centering
\includegraphics[scale=0.6]{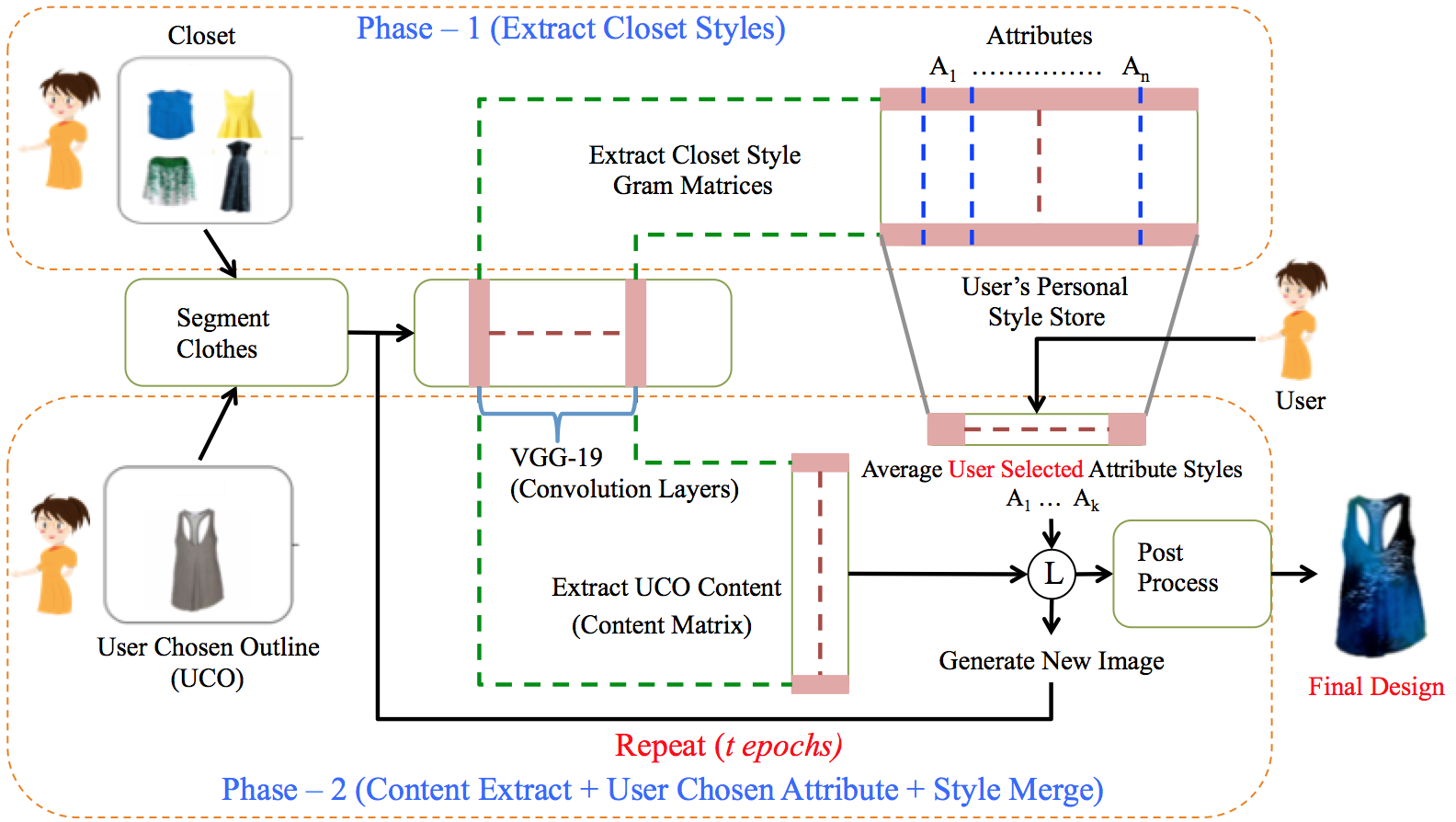}
\caption{Overall System Architecture. $A_1 ... A_n$ are all the attributes in the dataset \cite{liu2016deepfashion}, $A_1 ... A_k$ are set of attributes given by the user. $L$ is the total loss between gram matrix of modified (iteratively) UCO image \& gram matrices from user's personal style store (for $A_1 ... A_k$). In the first phase the user provides the system access to his / her closet images from where the user's fashion preferences are learned. In phase two, the user gives his / her choices (attributes such as \textit{Striped Top} or \textit{Chiffon}) with the desired outline of piece of clothing to get the new custom design.}
\label{fig:pipeline}
\end{figure*}

Although Convolutional Neural Networks provide state-of-the-art performance for multiple computer vision tasks, their complexity and opacity has been a substantial research question. Visualizing the features learned by the network has been addressed in multiple efforts. Zeilar \textit{et. al} \cite{DBLP:journals/corr/ZeilerF13} use a deconvolution network to reconstruct the features learned in each layer of the CNN. Simoyan \textit{et. al}. \cite{Simonyan14c} backpropagate the gradients generated for a class with respect to the input image to create an artificial image (the initial image is just random noise) that represents the class in the network. The separation of style and content in an image by Gatys \textit{et. al}. \cite{gatys2015texture} shows the \textit{variant} (content) and \textit{invariant} (style) parts of the image.

Our contribution in this paper is a pipeline to learn the user's unique fashion sense and generate new design patterns based on their preferences. Figure \ref{fig:sample} shows a sample clothing item generated using \textit{neural style transfer}. The first clothing item given by the user provides the shape for the new dress. The second is initially provided by the user from his/her closet to learn their preference. The third is the final generated design for the user (the generated sample contains styles from multiple pieces of the user's clothing).

The following sections discuss the related work, how neural style transfer works, our system architecture, experiments conducted and their results.

\section{Related Work}
Prior research on fashion data in the computer vision community has dealt with a whole range of challenges including clothes classification, predicting attributes of clothes and the retrieval of items \cite{kalantidis2013getting}\cite{bossard2012apparel}\cite{xiao2015learning}\cite{kiapour2014hipster}. Liu \textit{et. al} \cite{liu2016deepfashion} create a robust fashion dataset of about 800,000 images that contains annotations for various types of clothes, their attributes and the location of landmarks as well as cross-domains pairs. They also design a CNN to predict attibutes and landmarks. The architecture is based on a 16 layer VGGNet and adds convolution and fully-connected layers to train a network to predict them. Phillip \textit{et. al} \cite{pix2pix2016} perform image to image translation using a conditional adversarial network. They perform experiments to generate various fashion accessories when provided with a sketch of the item. 

We use a 19-layer pre-trained VGGNet \cite{Simonyan14c} that is trained on the imagenet dataset \cite{russakovsky2015imagenet}. The network consists of 8 convolutional layers and 3 fully-connected layers. It is trained to predict 1000 classes (from the Imagenet challenge). The network is known to be robust and the features generated have been used to solve multiple downstream tasks. Gatys \textit{et al.} use the pre-trained VGGNet to extract style and content features.

Johnson \textit{et. al}. \cite{Johnson2016Perceptual} create an image transformation network trained to transform the image with the given style. A feed-forward transformation network is trained to run real-time using perceptual loss functions that depend on high-level features from a pre-trained loss network rather than the per-pixel loss function based on low level pixel information. The trained network does not start transforming the image from white-noise but generates the output directly, thus speeding up the process.

Gatys \textit{et al.} \cite{gatys2015neural, gatys2016image} describes the process of using image representations encoded by multiple layers of VGGNet to separate the content and style of images and recombine them to form new images. The idea of style extraction is based on the texture synthesis process that represents the texture as a Gram Matrix of the feature maps generated from each convolutional layer. The style is extracted as a weighted set of gram matrices across all convolutional layers of the pre-trained VGGNet when it processes an image. The content is obtained from feature maps extracted from the higher layers of the network when the image is processed. The style and content losses are computed as the mean squared error (MSE) between the features maps and Gram matrices of the original image and a randomly generated image (initiated from white noise). Minimizing the loss transforms the white noise to a new artistic image. 

We use the method described above to generate new fashion designs.


\section{Preliminaries}
This section describes how the style and content is extracted from an image using \textit{neural style transfer} \cite{gatys2015neural}. We use the implementation given by \cite{Smith2016}, a pre-trained 19 layer VGGnet model (VGG-19) that takes a content image and a set of style images as input.


Consider an input image $x$ and convolutional neural network $NN$. Every convolution layer $l$ in the convolutional network has $N_l$ distinct filters. Upon completion of the convolution operation (and the activation function being applied), let the feature map computed have height $h$ and width $w$. The flattened map (into a single vector) has a size of $M_l = 1\times(hxw)$. Thus, the feature maps at every layer $l$ can be given as $F_{ij}^l \in \mathcal{R}^{N_l \times  M_l}$ where $F_{ik}^l$ represents the activation of the $i^{th}$ filter at position $k$. 


\subsection{Style Extraction}
The Gram matrix at layer $l$ is given by $G ^ l \in R ^ {N_l \times N_l}$ where $G_{ij} ^ l$ is calculated by the dot product of the feature maps $i$ and $j$ for layer $l$:
\begin{equation}
G_{ij} ^ l = \sum_k F_{ik}^l F_{jk}^l
\label{eq:gram_matrix}
\end{equation}

The dot product computes the similarities between feature maps. Thus the Gram matrix $G^l$ invariably contains image points that are consistent between the maps while inconsistent features become
0.

Consider two images $x$ (input image used to transfer the style) and $\hat{x}$ (a randomly generated image from white noise). Let their corresponding Gram matrices be $G^l$ and $\hat{G}^l$. The style loss function is then computed for every layer as the mean squared error (MSE) between $G^l$ and $\hat{G}^l$ as

\begin{equation}
E_l = \frac{1}{4M_l^2N_l^2} \sum_{i,j} (G_{ij}^l - \hat{G}^l_{ij})^2
\label{eq:style_loss}
\end{equation}

$E_l$ is the style loss.  


\subsection{Content Extraction}
The feature maps from the higher layers in the model give a representation of the image that is more biased towards the content \cite{gatys2015texture}. We use the feature representations of the \textit{conv\_4\_2} layer to extract content. Given the feature representations in layer $l$ of the original image $x$ and the generated white noise image $\hat{x}$ as $F^l$ and $\hat{F}^l$ respectively, we define the content loss as the mean squared difference between the two:

\begin{equation}
\mathcal{L}_{content}(x,\hat{x},l) = \frac{1}{2} \sum_{i,j} (F_{ij}^l - \hat{F}_{ij}^l)^2
\label{eq:content_loss}
\end{equation}

The derivative of this loss with respect to the feature map at layer $l$ gives the gradient used to minimize the loss:
\begin{equation}
\frac{\partial{\mathcal{L}_{content}}}{\partial{F_{ij}^l}} =  \begin{array}
    {lr}(F^l - \hat{F}^l)_{ij}, & \text{if } F_{ij}^l > 0\\
        0, & \text{if } F_{ij}^l < 0
\end{array}
\label{eq:content_loss_gradiant}
\end{equation}



\begin{figure*}[h]
    \centering
    \includegraphics[width=\textwidth]{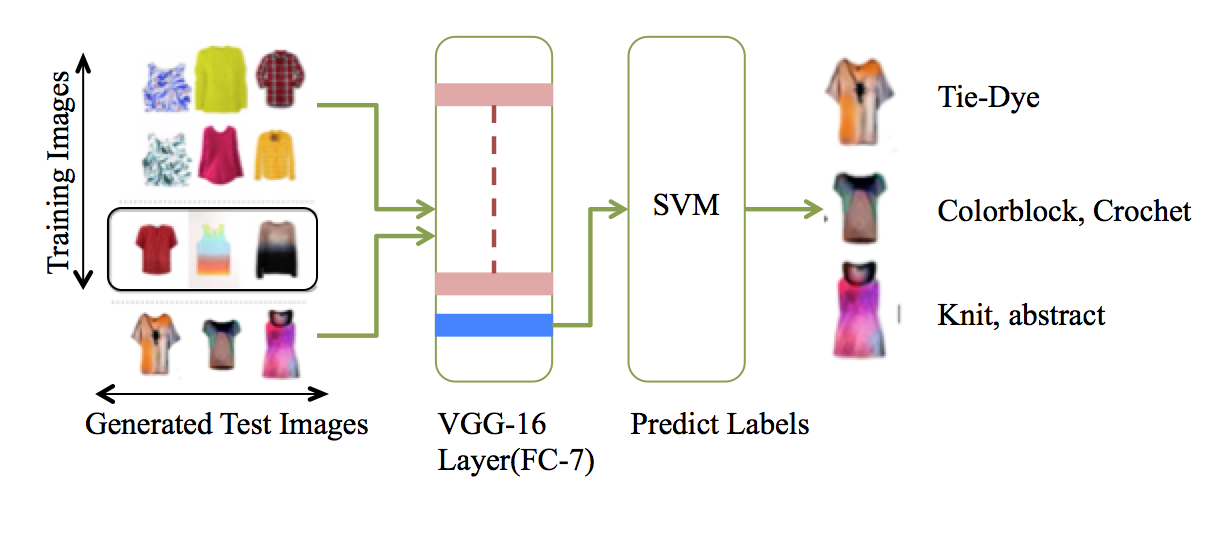}
    \caption{Evaluation model for predicting attribute labels on separate training and test generation images}
    \label{fig:separate_svm}
\end{figure*}

\section{System Architecture}
Figure \ref{fig:pipeline} shows the entire pipeline to personalize and design custom clothes for the user. There are four modules to the architecture, namely, preprocessing, personal style store creation, style transfer and post-processing to generate the final design. The following section discusses these modules in more detail.

To minimize the complexity of the problem, we consider images from the DeepFashion dataset \cite{liu2016deepfashion} that have a white background. The images contain only clothing objects with no humans or other artifacts. They are only upper-body or full-body apparel pieces.

\subsection{Preprocessing}
All images are resized to 512 x 512. The image is resized not by expanding/contracting the image, but by creating a temporary white background image of the above mentioned size. The original image is then placed at the center of that temporary image. This resizes the image to the expected size without deforming it. Also, the mask of the image is extracted and stored using the \textit{grabcut} utility \cite{rother2004grabcut}. This mask is used in the postprocessing step to get rid of patterns lying beyond the contours of the apparel. The attributes for the clothes are assumed to be provided and automatically labeling them is beyond the scope of this paper.

\subsection{Creating a Personal Style Store}
To learn the user's fashion preferences, the user initially provides the set of clothes from his / her closet. The Gram matrices $G_l$ (eq.\ref{eq:gram_matrix}) of all the clothes with their annotated attributes are calculated. Tensorflow \cite{tensorflow2015-whitepaper} allows us to get the partially computed functions $E_l$ in \ref{eq:style_loss} (where the gram matrices for $G_l$ are computed first and then $\hat{G}_l$ later). The style losses $E_l$ are thus stored in a dictionary with the associated attributes. A personal style store is constructed for each user. 

\subsection{Style Transfer}
To perform style transfer, two inputs are necessary. As shown in figure \ref{fig:pipeline}, the user inputs a list of attributes that he/she will like in their new garment. This list can be attributes like \textit{print} and \textit{stripes} or fabric such as \textit{chiffon}. In the current system, style is learned only for attribute types \textit{texture} and \textit{fabric}. The dress shape is not considered as a representation of the style of that object. Apart from these attributes, the user also gives an image that contains the shape of the dress they desire. This is called the User Chosen Outline (UCO). Let the attributes of the dresses in the closet be $A_1 ... A_n$. The selected user attributes are $A_1 ... A_k$ where $k << n$. The set of style loss functions having the corresponding attributes are selected from the user's personal style store. Although the style's extracted from the user's closet as a whole represent the his/her fashion sense, we pick the style functions of the chosen attributes because we assume the user's mental model of dress is likely to be similar to the styles extracted for those attributes. All selected functions are then combined to get a singular representation of the user's fashion choices.

For a style image $x$ and the initialized image $\hat{x}$, the style loss can be given as,
\begin{equation}
\mathcal{L}_{s}(\hat{x}, x) = \sum_{l=0}^L w_l E_l
\end{equation}
where $\mathcal{L}_{s}$ is the style loss for a single image. 

The combined loss is given by:
\begin{equation}
\mathcal{L}_{style} = \frac{1}{S}\sum_{s=0}^S{W_{s} \mathcal{L}_{s}}
\end{equation}
Here, $\mathcal{L}_{style}$ is the style loss computed over $S$ select functions.

The number of images for every attribute picked depends on the distribution of the particular attribute across the entire list of images present. The higher the frequency of the attribute in the distribution, the higher is the bias towards a certain label and suppresses the effect of the others. This makes certain image characteristics more pronounced in the final dress than others. Hence, to offset the bias the weight $W_{s}$ is utilized.

\textit{\textbf{Total Loss}} is the summation of the style and content losses obtained. 

\begin{equation}
\mathcal{L}_{total} = \alpha\mathcal{L}_{content}(C, x) + \beta\mathcal{L}_{style}(S, x)
\end{equation}
Here, $\alpha$ and $\beta$ are the weights assigned to the content and style losses respectively. C is the user chosen outline (UCO). An LBFGS optimizer is used to minimize the loss. The output image is then post-processed to get the final image. The objective is to minimize the content and style losses.

\begin{figure*}[h]
    \centering
    \includegraphics[scale=0.8]{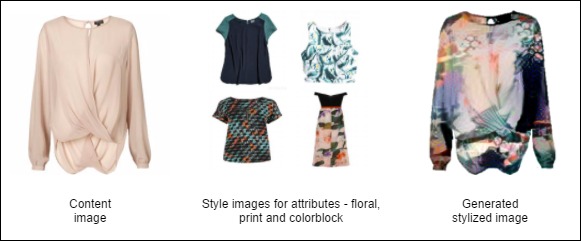}
    \caption{Multiple styles reinforced in a content image}
    \label{fig:multiple_styles_one_content}
\end{figure*}

\subsection{Postprocessing}
The output image contains patches of patterns transferred across the entire image. We resize the image to its original dimensions and apply the mask (of the UCO image) extracted to white out the background and get the transformed clothing object as the final resultant dress. 

\section{Experiments \& Results}
We present two approaches to evaluate the results of personalization using style transfer.


\subsection{Predicting Attribute labels}
Quantitative evaluation for personalization models is a challenging task. A standard approach is to create a survey of mechanical turk and ask users if the styles have been transferred properly and if the new dress designs are personalized given a wardrobe. But fashion presents a unique challenge as it is highly dependent on the user's taste for different kinds of clothing. Instead a different tact is applied.

Figure \ref{fig:separate_svm} shows how the evaluation is performed. We check if style is imparted on the given UCO image by verifying if the classifier is able to identify the style attributes present in it. An SVM is trained to learn attributes of the clothes present in the user's closet using the features generated from a 16-layer VGGNet (our system uses the 19 layer for fashioning the clothes). The test dataset is created by generated a random combination of attributes (these combinations are likely not present in the training image closet). For these random combinations of attributes, the new dress images are generated. Once featurized by a pre-trained VGG-16, we check the SVM's ability to predict the combinations of attributes.

\begin{figure}[h]
    \centering
    \includegraphics[scale=0.5]{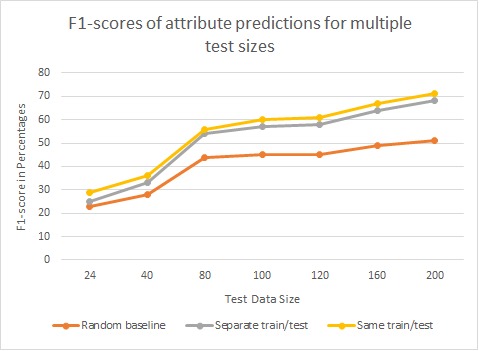}
    \caption{Bar-chart showing F1-scores for the baseline and our model on actual test data using separate training and test generation images, and using same images for training and test data generation}
    \label{fig:f1_scores_400}
\end{figure}

The UCO images and the set of images used for styling are maintained separately. There are a total of 400 UCOs and 100 images from the user's wardrobe. There are two kinds of tests considered in the experiment. In the first, the test images are generated from a set of images separate from the styles extracted from the training but with similar attributes. In the second, the test images are generated from the styles extracted from the training data itself. Figure \ref{fig:f1_scores_400} shows the F1-score for a varying number of test images generated. The consistent performance above the baseline suggests the style is likely transferred and the SVM is able the classify based on features generated. 

Our experiments with increasing the number of images used for gaining more styles showed a drop in the F1 score, suggesting that an increasing number of style functions impact the quality of the result, thus making it difficult to identify patterns. Hence it is necessary to limit the number of style functions used to generate the new dress.

\begin{figure*}[h]
    \centering
    \includegraphics[scale=0.6]{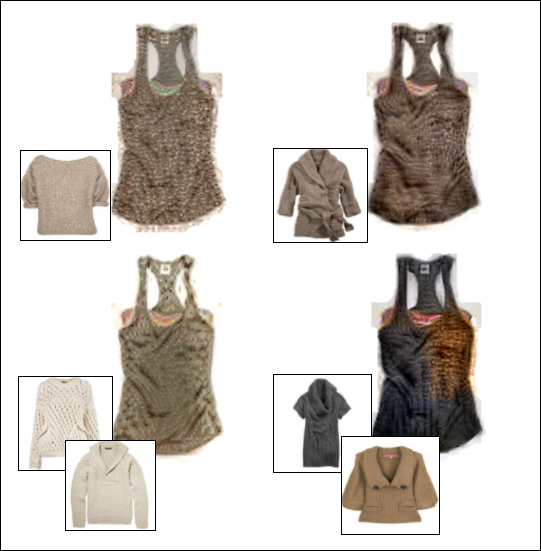}
    \caption{Styles extracted from multiple images for the same attribute "knit"}
    \label{fig:same_style_multiple_content}
\end{figure*}

\subsection{Qualitative  evaluation}
We analyze the quality of dress images by seeing how similar they are to the style images used in the personalization process. The quality of the generated image is impacted by a number of factors. The effect of various hyper-parameters is measured. The Figure \ref{fig:multiple_styles_one_content} shows an image of a \textit{sheer draped blouse} changed to adopt the styles extracted from a couple of images. The result is a nice blend of patterns borrowed from the style images given.

A single style superimposed on the same content image, but using multiple distinct style images, produces interesting results. Figure \ref{fig:same_style_multiple_content} presents the style of four different \textit{knit} garments over a \textit{tank top}. Four different textures of the same fabric produce distinct results.

\section{Conclusions \& Future Work}
In this paper, we show an initial pipeline to generate new designs for clothes based on the preference of the user. The results indicate that style transfer happens successfully and is personalized for the closet of a user. In the future we will like to improve the performance of the pipeline as it is time consuming to generate a new design. Also, we plan to experiment with better methods to personalize and generate designs with higher resolutions.

%
\bibliographystyle{abbrv}
\bibliography{sigproc}  

\begin{thebibliography}{10}

\bibitem{tensorflow2015-whitepaper}
M.~Abadi, A.~Agarwal, P.~Barham, E.~Brevdo, Z.~Chen, C.~Citro, G.~S. Corrado,
  A.~Davis, J.~Dean, M.~Devin, S.~Ghemawat, I.~Goodfellow, A.~Harp, G.~Irving,
  M.~Isard, Y.~Jia, R.~Jozefowicz, L.~Kaiser, M.~Kudlur, J.~Levenberg,
  D.~Man\'{e}, R.~Monga, S.~Moore, D.~Murray, C.~Olah, M.~Schuster, J.~Shlens,
  B.~Steiner, I.~Sutskever, K.~Talwar, P.~Tucker, V.~Vanhoucke, V.~Vasudevan,
  F.~Vi\'{e}gas, O.~Vinyals, P.~Warden, M.~Wattenberg, M.~Wicke, Y.~Yu, and
  X.~Zheng.
\newblock {TensorFlow}: Large-scale machine learning on heterogeneous systems,
  2015.
\newblock Software available from tensorflow.org.

\bibitem{bossard2012apparel}
L.~Bossard, M.~Dantone, C.~Leistner, C.~Wengert, T.~Quack, and L.~Van~Gool.
\newblock Apparel classification with style.
\newblock In {\em Asian conference on computer vision}, pages 321--335.
  Springer, 2012.

\bibitem{DBLP:journals/corr/ChenPK0Y16}
L.~Chen, G.~Papandreou, I.~Kokkinos, K.~Murphy, and A.~L. Yuille.
\newblock Deeplab: Semantic image segmentation with deep convolutional nets,
  atrous convolution, and fully connected crfs.
\newblock {\em CoRR}, abs/1606.00915, 2016.

\bibitem{efros2001image}
A.~A. Efros and W.~T. Freeman.
\newblock Image quilting for texture synthesis and transfer.
\newblock In {\em Proceedings of the 28th annual conference on Computer
  graphics and interactive techniques}, pages 341--346. ACM, 2001.

\bibitem{efros1999texture}
A.~A. Efros and T.~K. Leung.
\newblock Texture synthesis by non-parametric sampling.
\newblock In {\em Computer Vision, 1999. The Proceedings of the Seventh IEEE
  International Conference on}, volume~2, pages 1033--1038. IEEE, 1999.

\bibitem{gatys2015texture}
L.~Gatys, A.~S. Ecker, and M.~Bethge.
\newblock Texture synthesis using convolutional neural networks.
\newblock In {\em Advances in Neural Information Processing Systems}, pages
  262--270, 2015.

\bibitem{gatys2015neural}
L.~A. Gatys, A.~S. Ecker, and M.~Bethge.
\newblock A neural algorithm of artistic style.
\newblock {\em arXiv preprint arXiv:1508.06576}, 2015.

\bibitem{gatys2016image}
L.~A. Gatys, A.~S. Ecker, and M.~Bethge.
\newblock Image style transfer using convolutional neural networks.
\newblock In {\em Proceedings of the IEEE Conference on Computer Vision and
  Pattern Recognition}, pages 2414--2423, 2016.

\bibitem{AmazonCamera}
B.~Heater.
\newblock Amazon’s new echo look has a built-in camera for style selfies.
\newblock
  \url{https://techcrunch.com/2017/04/26/amazons-new-echo-look-has-a-built-in-camera-for-style-selfies/}.
\newblock Accessed: 2017-06-02.

\bibitem{heeger1995pyramid}
D.~J. Heeger and J.~R. Bergen.
\newblock Pyramid-based texture analysis/synthesis.
\newblock In {\em Proceedings of the 22nd annual conference on Computer
  graphics and interactive techniques}, pages 229--238. ACM, 1995.

\bibitem{pix2pix2016}
P.~Isola, J.-Y. Zhu, T.~Zhou, and A.~A. Efros.
\newblock Image-to-image translation with conditional adversarial networks.
\newblock {\em arxiv}, 2016.

\bibitem{Johnson2016Perceptual}
J.~Johnson, A.~Alahi, and L.~Fei-Fei.
\newblock Perceptual losses for real-time style transfer and super-resolution.
\newblock In {\em European Conference on Computer Vision}, 2016.

\bibitem{julesz1962visual}
B.~Julesz.
\newblock Visual pattern discrimination.
\newblock {\em IRE transactions on Information Theory}, 8(2):84--92, 1962.

\bibitem{kalantidis2013getting}
Y.~Kalantidis, L.~Kennedy, and L.-J. Li.
\newblock Getting the look: clothing recognition and segmentation for automatic
  product suggestions in everyday photos.
\newblock In {\em Proceedings of the 3rd ACM conference on International
  conference on multimedia retrieval}, pages 105--112. ACM, 2013.

\bibitem{kiapour2014hipster}
M.~H. Kiapour, K.~Yamaguchi, A.~C. Berg, and T.~L. Berg.
\newblock Hipster wars: Discovering elements of fashion styles.
\newblock In {\em European conference on computer vision}, pages 472--488.
  Springer, 2014.

\bibitem{krizhevsky2012imagenet}
A.~Krizhevsky, I.~Sutskever, and G.~E. Hinton.
\newblock Imagenet classification with deep convolutional neural networks.
\newblock In {\em Advances in neural information processing systems}, pages
  1097--1105, 2012.

\bibitem{kwatra2003graphcut}
V.~Kwatra, A.~Sch{\"o}dl, I.~Essa, G.~Turk, and A.~Bobick.
\newblock Graphcut textures: image and video synthesis using graph cuts.
\newblock In {\em ACM Transactions on Graphics (ToG)}, volume~22, pages
  277--286. ACM, 2003.

\bibitem{linden2003amazon}
G.~Linden, B.~Smith, and J.~York.
\newblock Amazon. com recommendations: Item-to-item collaborative filtering.
\newblock {\em IEEE Internet computing}, 7(1):76--80, 2003.

\bibitem{liu2016deepfashion}
Z.~Liu, P.~Luo, S.~Qiu, X.~Wang, and X.~Tang.
\newblock Deepfashion: Powering robust clothes recognition and retrieval with
  rich annotations.
\newblock In {\em Proceedings of the IEEE Conference on Computer Vision and
  Pattern Recognition}, pages 1096--1104, 2016.

\bibitem{portilla2000parametric}
J.~Portilla and E.~P. Simoncelli.
\newblock A parametric texture model based on joint statistics of complex
  wavelet coefficients.
\newblock {\em International journal of computer vision}, 40(1):49--70, 2000.

\bibitem{DBLP:journals/corr/RenHG015}
S.~Ren, K.~He, R.~B. Girshick, and J.~Sun.
\newblock Faster {R-CNN:} towards real-time object detection with region
  proposal networks.
\newblock {\em CoRR}, abs/1506.01497, 2015.

\bibitem{AmazonFashion}
J.~D. REY.
\newblock Amazon won a patent for an on-demand clothing manufacturing
  warehouse.
\newblock
  \url{https://www.recode.net/2017/4/18/15338984/amazon-on-demand-clothing-apparel-manufacturing-patent-warehouse-3d}.
\newblock Accessed: 2017-06-02.

\bibitem{rother2004grabcut}
C.~Rother, V.~Kolmogorov, and A.~Blake.
\newblock Grabcut: Interactive foreground extraction using iterated graph cuts.
\newblock In {\em ACM transactions on graphics (TOG)}, volume~23, pages
  309--314. ACM, 2004.

\bibitem{russakovsky2015imagenet}
O.~Russakovsky, J.~Deng, H.~Su, J.~Krause, S.~Satheesh, S.~Ma, Z.~Huang,
  A.~Karpathy, A.~Khosla, M.~Bernstein, et~al.
\newblock Imagenet large scale visual recognition challenge.
\newblock {\em International Journal of Computer Vision}, 115(3):211--252,
  2015.

\bibitem{Simonyan14c}
K.~Simonyan and A.~Zisserman.
\newblock Very deep convolutional networks for large-scale image recognition.
\newblock {\em CoRR}, abs/1409.1556, 2014.

\bibitem{Smith2016}
C.~Smith.
\newblock neural-style-tf.
\newblock \url{https://github.com/cysmith/neural-style-tf}, 2016.

\bibitem{trewin2000knowledge}
S.~Trewin.
\newblock Knowledge-based recommender systems.
\newblock {\em Encyclopedia of library and information science}, 69(Supplement
  32):180, 2000.

\bibitem{wei2000fast}
L.-Y. Wei and M.~Levoy.
\newblock Fast texture synthesis using tree-structured vector quantization.
\newblock In {\em Proceedings of the 27th annual conference on Computer
  graphics and interactive techniques}, pages 479--488. ACM
  Press/Addison-Wesley Publishing Co., 2000.

\bibitem{xiao2015learning}
T.~Xiao, T.~Xia, Y.~Yang, C.~Huang, and X.~Wang.
\newblock Learning from massive noisy labeled data for image classification.
\newblock In {\em Proceedings of the IEEE Conference on Computer Vision and
  Pattern Recognition}, pages 2691--2699, 2015.

\bibitem{DBLP:journals/corr/ZeilerF13}
M.~D. Zeiler and R.~Fergus.
\newblock Visualizing and understanding convolutional networks.
\newblock {\em CoRR}, abs/1311.2901, 2013.

\end{thebibliography}
%
%
\end{document}